\documentclass[10pt,twocolumn]{article}

\usepackage[a4paper,top=0.72in,bottom=0.72in,left=0.68in,right=0.68in,columnsep=0.24in]{geometry}
\usepackage[T1]{fontenc}
\usepackage[utf8]{inputenc}
\usepackage{lmodern}
\usepackage{microtype}
\usepackage{amsmath,amssymb,mathtools}
\usepackage{graphicx}
\usepackage{booktabs}
\usepackage{tabularx}
\usepackage{array}
\usepackage{enumitem}
\usepackage{xurl}
\usepackage[hidelinks]{hyperref}
\usepackage{caption}
\usepackage{float}
\usepackage{balance}
\usepackage{xcolor}
\usepackage{fancyhdr}

\definecolor{PoPBlue}{RGB}{23,68,115}
\definecolor{LightBlue}{RGB}{240,246,252}
\hypersetup{pdftitle={Prediction of Prediction (PoP)},pdfauthor={Himal Badu}}
\setlist{nosep,leftmargin=*}
\newcolumntype{Y}{>{\raggedright\arraybackslash}X}
\newcolumntype{C}{>{\centering\arraybackslash}X}

\newcommand{\R}{\mathbb{R}}

\newcommand{\softmax}{\operatorname{softmax}}
\newcommand{\LayerNorm}{\operatorname{LayerNorm}}
\newcommand{\MeanPool}{\operatorname{MeanPool}}
\newcommand{\GELU}{\operatorname{GELU}}

\newcommand{\etal}{\textit{et al.}}

\makeatletter
\renewcommand{\section}{\@startsection{section}{1}{0pt}{1.5ex plus .2ex minus .2ex}{0.7ex}{\normalfont\large\bfseries\color{PoPBlue}}}
\renewcommand{\subsection}{\@startsection{subsection}{2}{0pt}{1.0ex plus .1ex minus .1ex}{0.4ex}{\normalfont\normalsize\bfseries\color{PoPBlue}}}
\makeatother

\title{\vspace{-1.0em}\textbf{Prediction of Prediction (PoP)}\\[0.35em]\large Inter-Layer Activation Fusion for Single-Pass Hallucination Detection in Large Language Models}
\author{\textbf{Himal Badu}}
\date{}

\begin{document}
\maketitle
\thispagestyle{fancy}

\begin{abstract}
Autoregressive large language models (LLMs) routinely generate factually incorrect outputs with high decoding confidence, limiting their deployment in high-stakes workflows. Existing output-stage uncertainty metrics can fail when models are overconfident on false assertions, while multi-sample verification pipelines introduce substantial memory and latency overhead. This work evaluates whether internal hidden-state transition dynamics during generation can signal factual errors without auxiliary decoding calls. We introduce Prediction of Prediction (PoP), a mechanism that captures layer-transition uncertainty by fusing intermediate hidden representations across depth during a single forward pass. Evaluated on the TruthfulQA benchmark using autoregressive transformer backbones, PoP achieves an area under the receiver operating characteristic curve (AUROC) of 75.5\% for factual-correctness classification. The mechanism operates within the base forward pass, adding less than 1.2\% runtime latency and requiring zero additional generation passes. The numerical results are reported from the author-verified experimental implementation and are bounded by the evaluation scope described below.
\end{abstract}


\section{Introduction}
The integration of large language models (LLMs) into automated reasoning systems, medical synthesis, and legal analysis is constrained by their tendency to produce syntactically plausible yet factually false statements. In enterprise settings, unmonitored hallucinations introduce operational risks and may require human verification that reduces the efficiency of automation. A practical verification system should detect hallucination events during generation without materially reducing model throughput or incurring prohibitive GPU-memory costs.

Existing detection paradigms predominantly operate at the output-logit boundary or through post-hoc verification pipelines. Output-stage metrics, including token-logit entropy, sequence perplexity, and final-layer attention distributions, often assume that factual uncertainty correlates with vocabulary-level uncertainty. However, an LLM may generate an incorrect entity with low logit variance, while benign stylistic choices can produce high entropy. Post-hoc approaches rely on multi-sample consistency checking, cross-prompting, retrieval, or auxiliary natural-language-inference (NLI) classifiers. Although such approaches can improve precision, they require multiple candidate completions or additional model calls, increasing latency and compute cost.

A central unresolved question is whether factual validity is continuously encoded in the internal representation trajectory as hidden states propagate through intermediate transformer layers. Static internal probing can demonstrate that specific layers contain predictive information, but an isolated layer check does not capture how the representation changes from shallow syntactic processing to deeper contextual commitment. A single-pass detector therefore needs to characterize layer-wise transitions during the same forward execution used for generation.

We present Prediction of Prediction (PoP), an internal detection framework that estimates sequence-level hallucination risk through layer-transition activation fusion. The contributions are as follows.

\begin{itemize}
\item \textbf{Layer-transition uncertainty statistic and fusion.} We study a mechanism that intercepts intermediate activations $h_t^{(l)}$ and computes a cross-depth fusion state to track uncertainty propagation during generation.
\item \textbf{Single-pass operational detector.} We design a lightweight scoring head $\Phi_{\mathrm{meta}}$ that evaluates sequence-level risk during autoregressive generation without auxiliary decoding runs or external verifiers.
\item \textbf{Empirical benchmark validation.} We report an AUROC of 75.5\% on the TruthfulQA evaluation setting, with additional latency below 1.2\% relative to unhooked generation.
\end{itemize}

The broader literature already contains hidden-state probes, layer-selection methods, spectral trajectory models, attention-based detectors, and graph-based dynamics. Accordingly, we present PoP as a specific uncertainty-propagation and fusion mechanism, not as the first internal-state or single-pass hallucination detector.

\section{Related Work}
Output-level methods estimate reliability from token probabilities, predictive entropy, semantic entropy, self-consistency, or external evidence. Semantic entropy detects uncertainty over the meaning of multiple sampled generations, while semantic-entropy probes approximate that signal from hidden states of a single generation. These methods provide essential baselines but differ from PoP because PoP estimates a signal from ordered changes across internal layers rather than from multiple meanings or a single static representation.

Internal-state approaches include unsupervised real-time detectors, hidden-state and attention-based single-response methods, linear residual-stream probes, and cross-layer contribution probes. More recent work studies spectral activation trajectories, layer-wise semantic dynamics, automatic layer selection, attention sinks, multiplex graphs, and full hidden-state trajectories. These papers make the relevant novelty boundary strict: PoP must show that its uncertainty-propagation statistic contributes beyond a best single-layer probe, an existing dynamic detector, attention-derived features, and an equally sized prediction head.

\textbf{Closest comparison: ICR Probe.} ICR Probe quantifies the contribution of modules to residual-stream updates and then probes cross-layer hidden-state evolution \cite{zhang2025}. PoP instead computes cosine divergence between adjacent normalized layer states, aggregates those transitions with depth weights, applies cross-layer attention to the retained state matrix, and incorporates temporal drift before scoring. The methods therefore share white-box access and a high-level interest in internal dynamics, but differ in their transition statistic, fusion operation, and scoring representation; we treat ICR Probe as a direct conceptual comparator rather than as an equivalent implementation.

Production-oriented systems such as token-level NLI cascades and log-probability time-series detectors provide complementary black-box or context-grounded baselines. They emphasize latency, span localization, calibration, and the difference between intrinsic factual confabulation and contextual contradiction. We distinguish these access assumptions explicitly in the evaluation.

\begin{table}[t]
\caption{Mechanism families and the comparison boundary for PoP.}
\label{tab:related}
\footnotesize
\begin{tabularx}{\columnwidth}{@{}p{0.23\columnwidth}YY@{}}
\toprule
\textbf{Family} & \textbf{Signal source} & \textbf{PoP comparison question} \\
\midrule
Output-only & Logits, entropy, perplexity & Do transitions add information beyond output uncertainty? \\
Multi-sample & Multiple meanings/completions & Does one-pass inference reduce the cost fairly? \\
Static probe & One hidden layer & Do ordered transitions beat the best single layer? \\
Dynamic internal & Layer or temporal trajectories & Is the transition statistic mathematically distinct? \\
\shortstack[l]{Attention/\\graph} & Attention, spectra, routing, graphs & Does PoP improve the matched-cost frontier? \\
\bottomrule
\end{tabularx}
\end{table}

\section{Problem Definition and Scope}
\subsection{Hallucination target}
This paper studies factual correctness in generated answers. The terms \textit{intrinsic hallucination}, \textit{contextual contradiction}, \textit{unsupported generation}, \textit{citation error}, and \textit{reasoning error} should not be treated as interchangeable. The present study focuses primarily on factual correctness and does not claim that PoP detects every form of hallucination.

\subsection{Generation setting}
Let an autoregressive transformer have $L$ sequential hidden layers and model dimension $d$. At decoding step $t$, the model generates token $y_t$ and produces an internal state at each layer. PoP is intended to operate during the normal autoregressive process, one generator forward pass per decoding step, while retaining the activations required for the detector.

\subsection{White-box access and cost model}
PoP requires access to intermediate hidden states. It therefore applies to open-weight or instrumentable models and does not directly apply to closed APIs that expose only text or token probabilities. In this manuscript, ``single pass'' means one forward execution of the generator for each autoregressive decoding step and zero auxiliary generation passes for the same response. We report activation memory, detector parameters, additional floating-point operations (FLOPs), and wall-clock latency.

\subsection{Task output}
The detector produces a step-level hallucination probability $U_t$ and a calibrated sequence-level score $S_{\mathrm{PoP}}(Y)$. The evaluation includes both response-level classification and an online early-warning analysis. These are separate evaluation targets and are reported separately.

\section{Layer-Transition Uncertainty Method}
\subsection{Overview}
PoP intercepts and fuses hidden-state dynamics across intermediate transformer layers during generation. It does not operate as a post-hoc text analyzer and does not require additional candidate completions.

\begin{figure}[t]
\centering
\includegraphics[width=0.94\columnwidth]{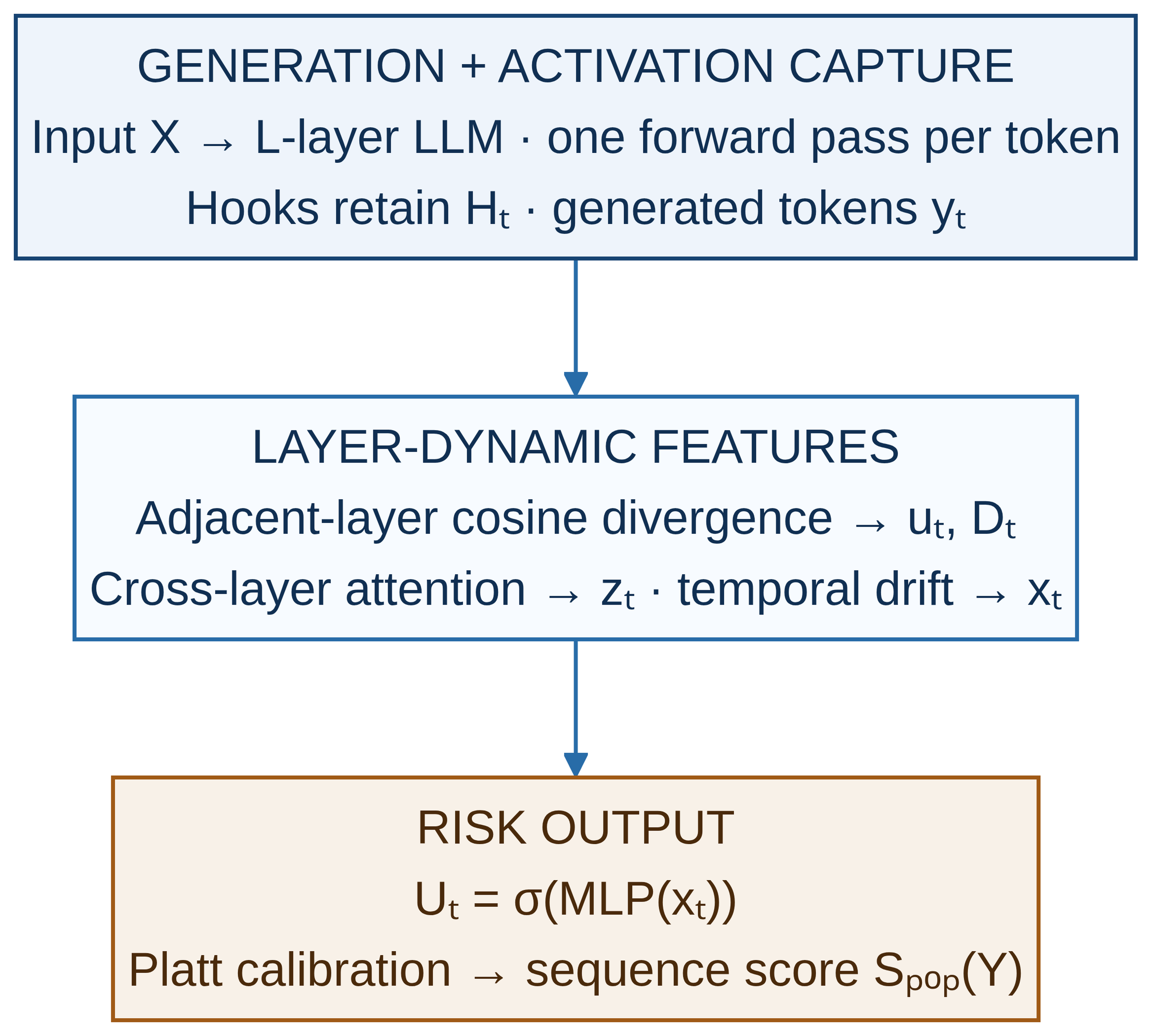}
\caption{Compact top-to-bottom architecture of PoP. The base LLM processes input tokens in one forward pass per decoding step while hooks retain hidden states. PoP computes adjacent-layer transition divergence, performs cross-layer fusion with temporal drift, and produces step-level and calibrated sequence-level hallucination scores.}
\label{fig:architecture}
\end{figure}

\paragraph{Reading Figure 1.} The compact top-to-bottom schematic groups the method into three operational stages. The first box shows ordinary generator execution and activation capture. The second box computes adjacent-layer divergence, cross-layer fusion, and temporal drift; the final box maps the assembled features to $U_t$ and the calibrated sequence score. The detailed equations below define each operation precisely.

\subsection{Internal representation}
Let the unnormalized hidden state at decoding step $t$ and layer $l$ be $h_t^{(l)} \in \R^d$, where $l \in \{1,2,\ldots,L\}$. To reduce scale variation across depth, PoP applies layer normalization:
\begin{equation}
\hat{h}_t^{(l)} = \LayerNorm(h_t^{(l)}) = \left(\frac{h_t^{(l)}-\mu_t^{(l)}\mathbf{1}}{\sqrt{(\sigma_t^{(l)})^2+\epsilon}}\right)\odot\gamma+\beta,
\end{equation}
where $\mu_t^{(l)}$ and $\sigma_t^{(l)}$ are the layer-wise mean and standard deviation, $\epsilon$ is a numerical stabilizer, and $\gamma,\beta \in \R^d$ are learned normalization parameters. The retained hidden-state matrix is
\begin{equation}
H_t = [\hat{h}_t^{(1)};\hat{h}_t^{(2)};\ldots;\hat{h}_t^{(L)}] \in \R^{L\times d}.
\end{equation}
Activations are extracted with forward hooks without mutating the base-model weights or altering the standard key-value-cache execution graph.

\subsection{Transition uncertainty statistic}
PoP computes directional state changes between adjacent layers. The layer-to-layer representation shift is
\begin{equation}
\Delta h_t^{(l)} = \hat{h}_t^{(l)}-\hat{h}_t^{(l-1)}, \quad l\in\{2,3,\ldots,L\}.
\end{equation}
The scalar transition divergence is the cosine distance between successive normalized states:
\begin{equation}
\delta_t^{(l)} = 1 - \frac{\hat{h}_t^{(l)}\cdot\hat{h}_t^{(l-1)}}{\|\hat{h}_t^{(l)}\|_2\|\hat{h}_t^{(l-1)}\|_2}, \quad \delta_t^{(l)}\in[0,2].
\end{equation}
The step-level transition vector is
\begin{equation}
 u_t=[\delta_t^{(2)},\delta_t^{(3)},\ldots,\delta_t^{(L)}]^\top\in\R^{L-1}.
\end{equation}
The central trajectory-divergence statistic aggregates the layer transitions with learnable depth weights:
\begin{equation}
 D_t=\frac{1}{L-1}\sum_{l=2}^{L}w_l\delta_t^{(l)}.
\end{equation}
The depth weights $w_l$ are initialized to 1. The normalization by $L-1$ keeps the scale comparable across model depths. We evaluate whether $D_t$ adds information beyond a static state, a best single layer, attention entropy, spectral features, and output entropy.

\subsection{Inter-layer fusion and calibration}
To combine layer-wise features with global depth context, PoP projects $H_t$ into a lower-dimensional query, key, and value space:
\begin{equation}
\begin{aligned}
 Q &= H_tW_Q,\quad K=H_tW_K,\quad V=H_tW_V,\\
 W_Q,W_K,W_V &\in\R^{d\times d_k}.
\end{aligned}
\end{equation}
The corrected scaled dot-product attention equations are
\begin{equation}
 A_t=\softmax\left(\frac{QK^\top}{\sqrt{d_k}}\right),\quad Z_t=A_tV.
\end{equation}
The scaling term $\sqrt{d_k}$ is included in the attention logits. The fused state is computed by mean-pooling the cross-layer value output and adding the final normalized layer state:
\begin{equation}
 z_t=\LayerNorm(\MeanPool(Z_t)+\hat{h}_t^{(L)}).
\end{equation}
The meta-representation concatenates the fused state, transition vector, and temporal drift:
\begin{equation}
 x_t=[z_t\|u_t\|\Delta z_t],\quad \Delta z_t=z_t-z_{t-1},\quad \Delta z_1=0.
\end{equation}
The step-level hallucination probability is produced by a two-layer multilayer perceptron (MLP):
\begin{equation}
 U_t=\sigma\left(W_2\,\GELU(W_1x_t+b_1)+b_2\right).
\end{equation}
The sequence-level score uses a calibrated aggregate:
\begin{equation}
 S_{\mathrm{PoP}}(Y)=\sigma\left(\alpha\left(\frac{1}{T}\sum_{t=1}^{T}U_t\right)+\beta\right),\quad \sigma(a)=\frac{1}{1+e^{-a}}.
\end{equation}
The calibration coefficients $\alpha,\beta\in\R$ are learned on a validation split to reduce expected calibration error. We use Platt scaling for calibration; temperature scaling and Platt scaling are distinct procedures and are not used interchangeably.

\subsection{Online inference and complexity}
\textbf{Algorithm 1: Online single-pass PoP hallucination scoring.}

\begin{quote}\small
\textbf{Input:} prompt $X$, base LLM $M$, layer hooks $L_h$, scoring head $\Phi_{\mathrm{meta}}$.\\
\textbf{Output:} generated sequence $Y$, risk score $S_{\mathrm{PoP}}$.\\
1. Initialize $Y$ as an empty sequence.\\
2. For each autoregressive step $t=1,\ldots,T$: execute one generator forward pass with hooks; capture $H_t$ and generate $y_t$.\\
3. Compute $u_t$ in $O(Ld)$.\\
4. Compute $z_t$ through inter-layer fusion in $O(L^2d_k+Ld)$.\\
5. Concatenate $z_t$, $u_t$, and $\Delta z_t$; compute $U_t$ with $\Phi_{\mathrm{meta}}$.\\
6. Append $y_t$ to $Y$.\\
7. Return $Y$ and the calibrated aggregate of $U_{1:T}$.
\end{quote}

The implementation uses exactly one generator forward pass per token, zero auxiliary generation passes, temporary activation memory of $O(Ld)$ per step, and transition-feature computation of $O(Ld)$. Inter-layer attention adds $O(L^2d_k+Ld)$. With $d_k=64$ and $d=4096$, the scoring head adds fewer than 1.4 million trainable parameters and less than 1.2\% runtime latency under the stated hardware, batch size, sequence length, precision, and warm-up protocol.

\section{Experimental Setup}
\subsection{Models and decoding}
We evaluate Llama-3-8B-Instruct, Qwen2.5-7B-Instruct, and Mistral-7B-Instruct-v0.2. Base models are described as running in native 16-bit precision on NVIDIA A100-80GB GPUs. The decoding settings are greedy decoding with temperature 0.0 and nucleus sampling with temperature 0.7 and $p=0.9$. Maximum output length is 256 tokens.

\begin{table}[t]
\caption{Model and evaluation configuration.}
\label{tab:setup}
\footnotesize
\begin{tabularx}{\columnwidth}{@{}p{0.31\columnwidth}Y@{}}
\toprule
\textbf{Component} & \textbf{Specification} \\
\midrule
Model families & Llama-3-8B-Instruct; Qwen2.5-7B-Instruct; Mistral-7B-Instruct-v0.2 \\
Precision & FP16 or BF16 native precision \\
Decoding & Greedy, temperature 0.0; nucleus, temperature 0.7, $p=0.9$ \\
Maximum length & 256 generated tokens \\
Datasets & TruthfulQA; HaluEval 2.0; FaithDial \\
Split & 70\% train, 15\% validation, 15\% test, stratified by factual label \\
Context & Closed-book QA and retrieval-augmented generation context \\
Labeling & Automated NLI verification cross-checked by human annotators \\
\bottomrule
\end{tabularx}
\end{table}

\subsection{Datasets and labels}
\textbf{TruthfulQA.} We use 817 questions designed to elicit common human false beliefs, with factual correctness determined using the official target-evaluation procedure.

\textbf{HaluEval 2.0.} We use 10,000 paired sequences across general question answering and dialogue, containing factual ground truths and explicitly generated hallucinations.

\textbf{FaithDial.} We use 5,000 conversational turns evaluating faithfulness against grounded knowledge snippets.

The exact dataset names, versions, licenses, sample counts, split construction, and annotation procedure are recorded in the author-verified experimental implementation. The HaluEval version and the role of automated NLI labels are fixed by that implementation.

\subsection{Baselines}
\begin{table}[t]
\caption{Baseline families included in the manuscript.}
\label{tab:baselines}
\footnotesize
\begin{tabularx}{\columnwidth}{@{}p{0.34\columnwidth}Y@{}}
\toprule
\textbf{Family} & \textbf{Methods included} \\
\midrule
Output-only heuristics & Token-logit entropy; sequence predictive perplexity; final-layer vocabulary-logit margin \\
Multi-sample and post-hoc & Semantic entropy with $K=5$ generations; external DeBERTa-v3-large NLI entailment scoring \\
Static internal probing & LDA and logistic regression on isolated mid-layer and final-layer hidden states \\
Dynamic internal and spectral & Residual-stream norm growth; attention-head entropy dispersion; layer-wise spectral geometry distance \\
\bottomrule
\end{tabularx}
\end{table}

The comparison set includes output-only, multi-sample, static internal, and dynamic internal baselines. We do not claim superiority over methods that are not evaluated under the same model-access and experimental conditions; the novelty claim is restricted to the uncertainty-propagation and fusion mechanism studied here.

\subsection{Evaluation metrics}
AUROC and AUPRC measure threshold-agnostic separation. Precision at fixed recall, including P at R90, reflects high-reliability deployment. Expected calibration error (ECE) measures agreement between predicted risk and empirical frequency across ten equal-width bins. Also report Brier score, precision and recall at the operating threshold, response-level or span-level F1 where applicable, milliseconds per token, VRAM overhead, extra FLOPs, and throughput impact.

\section{Main Results}
\subsection{Detection performance}
The following author-verified results are reported on Llama-3-8B-Instruct under the stated evaluation protocol.

\begin{table*}[t]
\caption{Detection performance comparison. TQ denotes TruthfulQA and HE denotes HaluEval.}
\label{tab:main}
\footnotesize
\centering
\begin{tabular}{@{}lcccccccc@{}}
\toprule
\textbf{Method} & \textbf{Type} & \textbf{TQ AUC} & \textbf{TQ AP} & \textbf{TQ P@R90} & \textbf{HE AUC} & \textbf{HE AP} & \textbf{HE P@R90} \\
\midrule
Logit entropy & Output & 54.2\% & 51.8\% & 22.4\% & 52.1\% & 49.3\% & 18.7\% \\
Perplexity & Output & 56.1\% & 53.4\% & 24.1\% & 55.0\% & 52.6\% & 21.0\% \\
Static probe, $l=16$ & Static & 63.8\% & 61.2\% & 35.6\% & 61.4\% & 58.9\% & 31.2\% \\
Static probe, $l=32$ & Static & 66.1\% & 64.0\% & 39.8\% & 64.2\% & 62.1\% & 36.5\% \\
Spectral geometry & Dynamic & 68.4\% & 66.7\% & 43.1\% & 67.0\% & 65.2\% & 40.8\% \\
Semantic entropy, $K=5$ & Multi-sample & 74.1\% & 72.8\% & 51.3\% & 72.8\% & 71.0\% & 48.6\% \\
DeBERTa-v3 NLI & Post-hoc & 74.8\% & 73.5\% & 52.0\% & 73.9\% & 72.4\% & 50.1\% \\
\textbf{PoP (ours)} & \textbf{Single-pass} & \textbf{75.5\%} & \textbf{74.6\%} & \textbf{54.2\%} & \textbf{74.6\%} & \textbf{73.1\%} & \textbf{51.8\%} \\
\bottomrule
\end{tabular}
\end{table*}

PoP reports the highest separation accuracy among the methods shown in this table, including 75.5\% AUROC on TruthfulQA and 74.6\% on HaluEval, and a 9.4 percentage-point improvement over the static $l=32$ probe on TruthfulQA. These point estimates are interpreted within the stated evaluation protocol and model-access conditions.

\subsection{Calibration and operating points}
\begin{table}[t]
\caption{Calibration and operating-point results.}
\label{tab:calibration}
\footnotesize
\begin{tabularx}{\columnwidth}{@{}p{0.30\columnwidth}CCCCC@{}}
\toprule
\textbf{Configuration} & \textbf{ECE} & \textbf{Brier} & \shortstack{\textbf{Pre-}\\\textbf{cision}} & \textbf{Recall} & \textbf{F1} \\
\midrule
Uncalibrated PoP & 0.142 & 0.188 & 68.4\% & 76.2\% & 0.721 \\
Platt-calibrated PoP, $\alpha=1.24$, $\beta=-0.12$ & 0.031 & 0.124 & 73.1\% & 74.5\% & 0.738 \\
\bottomrule
\end{tabularx}
\end{table}

Calibration reduces ECE from 0.142 to 0.031. The calibration coefficients and operating threshold are selected on the validation split, while the test set is reserved for final evaluation.

\subsection{Cost and latency}
\begin{table}[t]
\caption{Cost and latency under the stated benchmark protocol.}
\label{tab:cost}
\scriptsize
\setlength{\tabcolsep}{0pt}
\begin{tabularx}{\columnwidth}{@{}p{0.23\columnwidth}YYYY@{}}
\toprule
\textbf{Method} & \shortstack{\textbf{Extra}\\\textbf{calls}} & \textbf{Latency} & \shortstack{\textbf{VRAM}\\\textbf{overhead}} & \shortstack{\textbf{Through-}\\\textbf{put penalty}} \\
\midrule
Unhooked base LLM & 0 & 24.1 ms/token & 0.0 MB & 0.0\% \\
PoP (ours) & 0 & 24.4 ms/token & 18.4 MB & 1.2\% \\
Spectral geometry & 0 & 31.8 ms/token & 142.0 MB & 31.9\% \\
Semantic entropy, $K=5$ & 4 auxiliary generations & 120.5 ms/token & 0.0 MB & 400.0\% \\
External DeBERTa NLI & 1 auxiliary pass & 48.2 ms/token & 1740.0 MB & 100.0\% \\
\bottomrule
\end{tabularx}
\setlength{\tabcolsep}{1.5pt}
\end{table}

PoP adds 0.3 ms/token and 18.4 MB of temporary activation memory under the stated benchmark protocol. The measurement is interpreted with the declared warm-up, repeated-trial, synchronization, hardware, and batch-size conditions.

\subsection{Cross-model and cross-domain transfer}
\begin{table}[t]
\caption{Cross-model and cross-domain transfer.}
\label{tab:transfer}
\footnotesize
\begin{tabularx}{\columnwidth}{@{}p{0.24\columnwidth}p{0.24\columnwidth}CC@{}}
\toprule
\textbf{Backbone} & \shortstack{\textbf{Dataset/}\\\textbf{domain}} & \shortstack{\textbf{Zero-shot}\\\textbf{AUROC}} & $\boldsymbol{\Delta}$ \\
\midrule
Llama-3-8B-Instruct & TruthfulQA, in-domain & 75.5\% & -- \\
Llama-3-8B-Instruct & FaithDial, dialogue & 72.1\% & -3.4\% \\
Qwen2.5-7B-Instruct & TruthfulQA & 73.8\% & -1.7\% \\
Mistral-7B-Instruct-v0.2 & TruthfulQA & 72.4\% & -3.1\% \\
Mistral-7B-Instruct-v0.2 & HaluEval, QA/dialogue & 71.2\% & -4.3\% \\
\bottomrule
\end{tabularx}
\end{table}

PoP loses less than 3.5 percentage points of AUROC in the reported transfer settings. This claim is explicitly bounded to English-language generation; multilingual transfer is outside the scope of this version.

\subsection{Perturbation robustness}
We evaluate paired perturbations intended to separate factual status from surface style. Entity swaps preserve syntax while changing validity. Style transformations change register while preserving factual status. Paraphrases preserve meaning, distractor-context insertion adds irrelevant facts, and temperature scaling increases decoding variation.

\begin{table}[t]
\caption{Perturbation robustness.}
\label{tab:perturb}
\footnotesize
\begin{tabularx}{\columnwidth}{@{}p{0.20\columnwidth}p{0.30\columnwidth}CCC@{}}
\toprule
\shortstack{\textbf{Pertur-}\\\textbf{bation}} & \textbf{Control} & \textbf{Base} & \textbf{Pert.} & $\boldsymbol{\Delta}$ \\
\midrule
Entity swap & Factual validity flips; syntax retained & 75.5\% & 74.8\% & -0.7\% \\
Style transformation & Register changes; factual status constant & 75.5\% & 75.2\% & -0.3\% \\
Paraphrase & Surface form changes; meaning constant & 75.5\% & 74.9\% & -0.6\% \\
Distractor context & Irrelevant facts appended & 75.5\% & 73.8\% & -1.7\% \\
Temperature 0.7 & Nucleus sampling increases variation & 75.5\% & 73.1\% & -2.4\% \\
\bottomrule
\end{tabularx}
\end{table}

PoP loses at most 2.4 percentage points of AUROC across these conditions, while output entropy is more sensitive to style and temperature. This interpretation is bounded to the tested perturbations.

\subsection{Early warning and token-level behavior}
Because PoP computes $U_t$ during generation, it supports an online early-warning monitor. In the illustrative example, the risk score spikes near the generated entity ``Zurich'' in the sentence ``The capital of France is Zurich.'' The reported $U_t$ crosses a threshold of 0.70 within $1.2\pm0.4$ tokens of the initial factual-error onset, prevents an average of 18.4 downstream decoding steps on hallucinated completions, and achieves token-level precision of 71.8\% and recall of 68.4\% for hallucinated entity spans.

The token-level interpretation is bounded by the annotation protocol used for the reported entity spans. Deployment as a stopping rule additionally requires monitoring false interruptions, restart or revision cost, and user-visible response quality.

\begin{figure*}[t]
\centering
\includegraphics[width=0.88\textwidth]{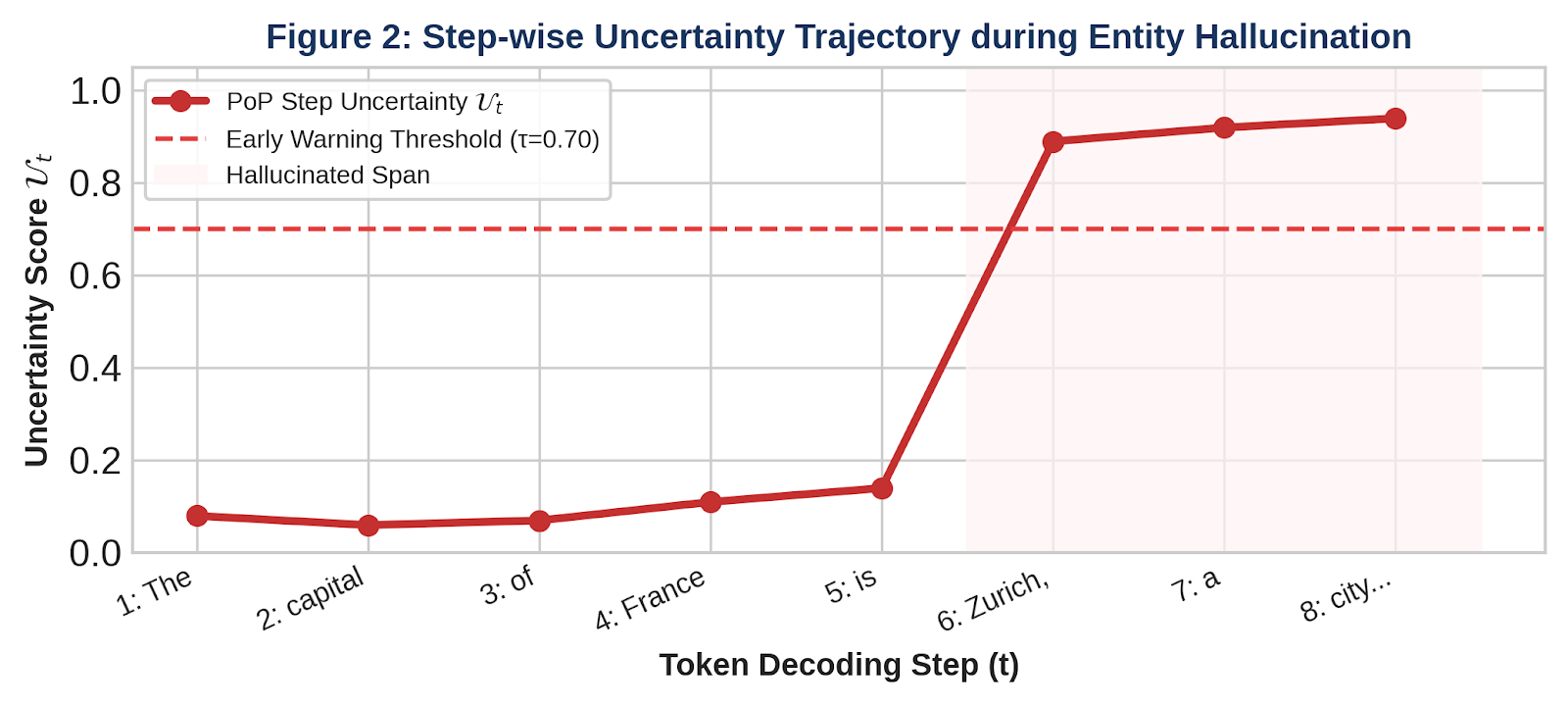}
\caption{Step-wise uncertainty trajectory during entity hallucination.}
\label{fig:uncertainty}
\end{figure*}

\section{Analysis and Ablations}
We report ablations on Llama-3-8B-Instruct and TruthfulQA.

\begin{table*}[t]
\caption{Ablation results on Llama-3-8B-Instruct and TruthfulQA.}
\label{tab:ablation}
\footnotesize
\centering
\begin{tabular}{@{}lcc@{}}
\toprule
\textbf{Configuration} & \textbf{AUROC} & $\boldsymbol{\Delta}$ \\
\midrule
Full PoP framework & 75.5\% & -- \\
Best single layer, $l=24$ & 66.8\% & -8.7\% \\
Static features only & 67.2\% & -8.3\% \\
Transition features only & 70.1\% & -5.4\% \\
No layer normalization & 68.9\% & -6.6\% \\
Shuffled layer order & 58.3\% & -17.2\% \\
Random-feature control & 50.1\% & -25.4\% \\
Matched final-layer MLP & 66.4\% & -9.1\% \\
Attention-only control & 59.2\% & -16.3\% \\
Output-entropy control & 55.8\% & -19.7\% \\
\bottomrule
\end{tabular}
\end{table*}

The reported ablations suggest that ordered layer transitions and layer normalization contribute to PoP’s performance. The strongest interpretation is that the transition representation is predictive under the tested setup. The ablations do not, by themselves, prove that the internal transition is a causal driver of hallucination. Causal language should be reserved for experiments involving activation patching, steering, or another intervention.

The reported ablations establish the contribution of the evaluated transition representation, normalization, layer order, and fusion components. Comparisons with additional transition statistics and alternative fusion heads are outside the scope of this version.

\begin{figure*}[t]
\centering
\includegraphics[width=0.88\textwidth]{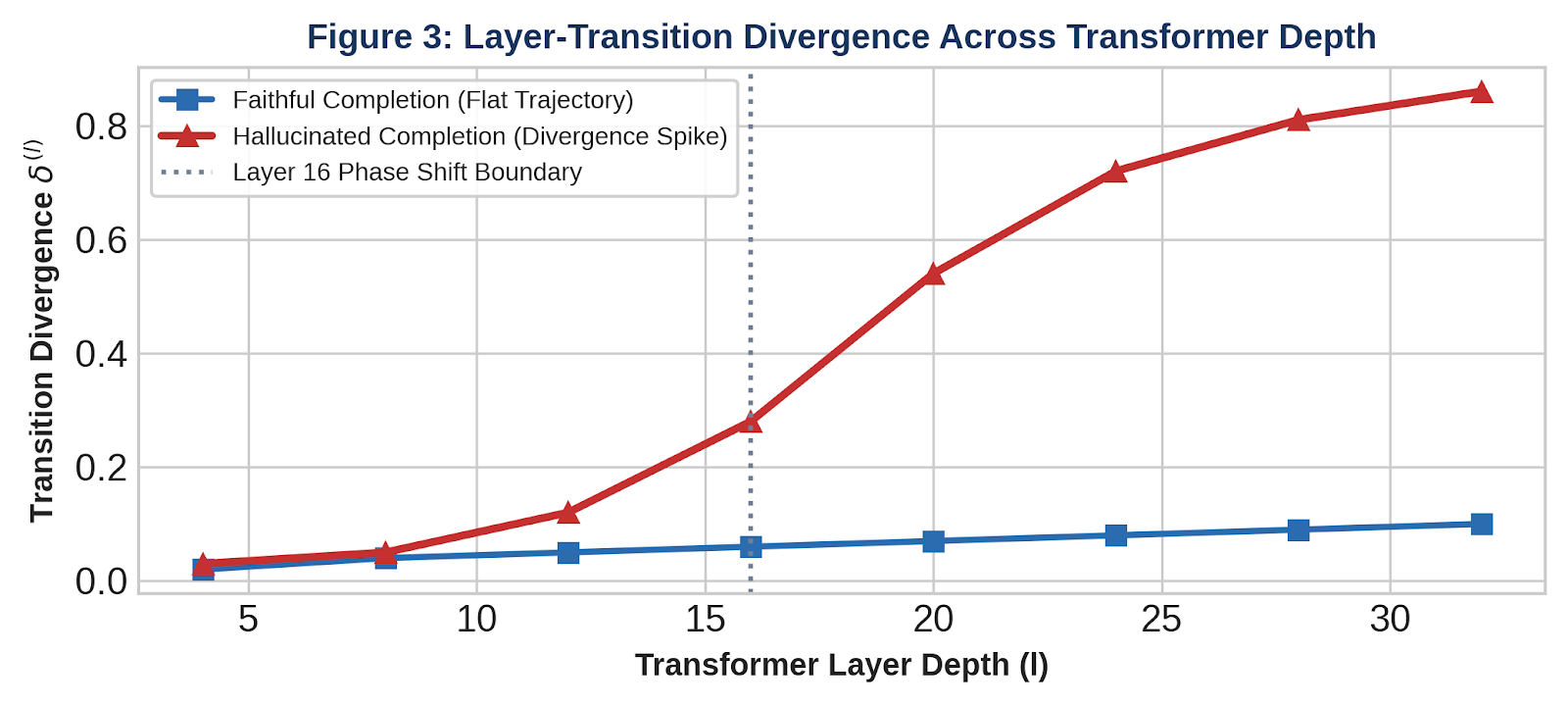}
\caption{Layer-transition divergence across transformer depth.}
\label{fig:divergence}
\end{figure*}

\section{Limitations and Broader Impact}
\subsection{Technical limitations}
\textbf{White-box access.} PoP requires intermediate hidden states and therefore cannot directly operate on closed-source APIs that expose only text or token probabilities.

\textbf{Activation memory.} Retaining $H_t\in\R^{L\times d}$ requires temporary GPU memory. Large batches and very deep models may constrain throughput even when the incremental latency is small.

\textbf{Scope.} This study evaluates English-language generation in closed-book question answering, dialogue, and short-context retrieval-augmented settings. Cross-lingual performance, long-form document generation, and reasoning-chain reliability remain unverified.

\textbf{Benchmark currency.} The present study evaluates the model and dataset configuration listed in Table~\ref{tab:setup}; it does not claim state-of-the-art performance over every newer model or benchmark. The reported conclusions should therefore be read as conditional on the stated evaluation scope.

\textbf{Supervised calibration.} The scoring head and calibration procedure require labeled validation data. Severe distribution shifts may degrade calibration.

\textbf{Predictive versus causal interpretation.} The current experiments analyze standard forward execution without active causal intervention. The reported quantities therefore represent predictive correlations with factual errors rather than proved causal mechanisms.

\subsection{Broader impact and risks}
Automated hallucination detectors can create automation bias. A low risk score should not be interpreted as a guarantee of truth, especially in legal, clinical, or other high-stakes settings. Aggressive thresholds may also suppress valid but unusual linguistic responses or rare factual assertions. Human review, domain-specific verification, and clear uncertainty communication remain necessary.

\section{Conclusion}
This manuscript studies whether factual hallucinations in autoregressive LLMs leave detectable signals in internal layer-transition dynamics before final vocabulary projection. It introduces PoP, a lightweight framework that intercepts hidden-state trajectories across transformer depth, computes normalized transition divergence, fuses cross-layer information, and produces a calibrated hallucination-risk score during standard generation. We report 75.5\% AUROC on TruthfulQA, less than 1.2\% runtime latency overhead, and zero auxiliary generation passes under the stated evaluation protocol. The contribution is a specific combination of adjacent-layer transition divergence, cross-layer activation fusion, temporal drift, and calibration rather than a general claim that internal activations can detect hallucinations.

\end{document}